\documentclass[12pt]{article}

\usepackage{sbc-template}
\usepackage{graphicx,url}
\usepackage[utf8]{inputenc}

\usepackage{lipsum}

\newcommand\blfootnote[1]{%
  \begingroup
  \renewcommand\thefootnote{}\footnote{#1}%
  \addtocounter{footnote}{-1}%
  \endgroup
}

\renewcommand{\refname}{Referências}
     
\title{The Brazilian Data at Risk in the Age of AI?}

\author{Raoni F. da S. Teixeira\inst{1}*,  Rafael B. Januzi\inst{2}, Fabio A. Faria\inst{2}*}

\address{Faculdade de Engenharia, Universidade Federal de Mato Grosso (FaEng-UFMT)\\
  Cuiab\'{a} -- MT -- Brasil \\ \texttt{raoni.teixeira@ufmt.br}
\nextinstitute
  Instituto de Ciência e Tecnologia, Universidade Federal de S\~{a}o Paulo (ICT-UNIFESP) \\ 
   S\~{a}o Jos\'{e} dos Campos -- SP -- Brasil \\ 
   \texttt{rafael.januzi@unifesp.br, ffaria@unifesp.br}
}

\begin{document} 

\maketitle

\begin{abstract}
  Advances in image processing and analysis as well as machine learning techniques have contributed to the use of biometric recognition systems in daily people tasks. These tasks range from simple access to mobile devices to tagging friends in photos shared on social networks and complex financial operations on self-service devices for banking transactions. On July 05th 2021, the Brazilian government announced acquisition of a biometric recognition system to be used nationwide. In the opposite direction to China, Europe and some American cities have already started the discussion about the legality of using biometric systems in public places, even banning this practice in their territory. In order to open a deeper discussion about the risks and legality of using these systems, this work exposes the vulnerabilities of biometric recognition systems, focusing its efforts on the face modality. Furthermore, it shows how it is possible to fool a biometric system through a well-known presentation attack approach in the literature called morphing. Finally, a list of ten concerns was created to start the discussion about the security of citizen data and data privacy law in the Age of Artificial Intelligence (AI).
\end{abstract}
     
\begin{resumo} 
  Com os avanços das técnicas de processamento e análise de imagens, o uso de sistemas de reconhecimento biométrico em tarefas cotidianas das pessoas já é uma realidade. Dentre essas tarefas estão desde um simples acesso aos dispositivos móveis até a marcação de amigos em fotos compartilhadas em redes sociais e as complexas operações financeiras em equipamentos de autoatendimento para transações bancárias. Em 5 de julho de 2021, o governo brasileiro anunciou a compra de um sistema de reconhecimento biométrico para ser utilizado em todo território nacional. Neste sentido, este artigo propõe a abertura de uma discussão mais aprofundada sobre a adoção de tais sistemas para a identificação dos cidadãos brasileiros e quais os problemas que podem emergir se o sistema não for bem projetado, implantado e gerenciado. Além disso, uma lista de dez questões foi criada para iniciar essa conversa sobre segurança dos dados dos brasileiros na Era da Inteligência Artificial (IA) e o respeito à Lei Geral de Proteção dos Dados (LGPD)\blfootnote{\textbf{*R. F. da S. Teixeira e F. A. Faria contribuiram igualmente para este trabalho.}}.
\end{resumo}


\section{Introdução}

Os significativos avanços em algoritmos de processamento e análise de imagens digitais têm contribuído para que os sistemas de reconhecimento facial (RF) rapidamente se tornassem um componente importante na rotina de grande parcela da população mundial. Essas tecnologias fazem parte da vida das pessoas desde o simples desbloqueio de seus dispositivos móveis para acesso de suas contas à marcação de amigos em postagens compartilhadas nas redes sociais (e.g., Facebook e Instagram) ~\cite{morphing_survey_2020}.  

A adoção generalizada de técnicas de reconhecimento facial tem levantado uma série de preocupações relacionadas com uso dessas tecnologias, principalmente em espaços públicos. 
Uma das preocupações é que essas tecnologias venham a ser utilizadas para restrigir liberdades e monitorar e/ou perseguir grupos minoritários e políticos. 
Organizações como a Comissão Francesa para Proteção de Dados Pessoais (\textit{Commission Nationale de l'Informatique et des Libertés} -- CNIL)~\cite{CNILreport} e o Instituto AINow da Universidade de Nova York~\cite{AINOWreport} têm levantado essas preocupações chamando atenção da opinião pública para a importância do tema e a urgência do debate.

Um efeito prático dessas discussões é que várias cidades americanas tais como Minneapolis\footnote{{https://techcrunch.com/2021/02/12/minneapolis-facial-recognition-ban/ Acessados em 20/07/2021.}}, São Francisco\footnote{{https://www.nytimes.com/2019/05/14/us/facial-recognition-ban-san-francisco.html }} e Oakland\footnote{{https://www.sfchronicle.com/bayarea/article/Oakland-bans-use-of-facial-recognition-14101253.php }} já proibiram o uso de sistema  de reconhecimento facial em serviços municipais, incluindo forças policiais. 
De encontro com essas cidades americanas, em abril de 2021, a Comissão Européia propôs um documento regulatório do uso de tecnologias de inteligência artificial e dentre as aplicações consideradas de alto risco está sistema de reconhecimento biométrico e  seu uso em espaços acessíveis ao público para fins de aplicação da lei é proibido a princípio\footnote{https://digital-strategy.ec.europa.eu/en/policies/regulatory-framework-ai}. 

Já no caminho oposto do mundo está o Brasil que acaba de anunciar a compra de um sistema nacional de reconhecimento biométrico por meio do consórcio Iafis Brasil e uma empresa fornecedora\footnote{https://www.telesintese.com.br/policia-federal-adquire-sistema-de-identificacao-biometrica-com-reconhecimento-facial/}\footnote{https://www.gov.br/pf/pt-br/assuntos/noticias/2021/07/policia-federal-implementa-nova-solucao-automatizada-de-identificacao-biometrica}. Acredita-se que a implantação de um sistema dessa importância precisaria de algum procedimento deliberativo com objetivo de ouvir todas as partes interessadas e abrir uma discussão em profundidade sobre os impactos de seu uso\footnote{{https://www.telesintese.com.br/sistema-de-reconhecimento-facial-da-pf-traz-riscos-e-fere-a-constituicao/ }}. {Essa discussão deve ocorrer observando inclusive o arcabouço legal instituído pela Lei Geral de Proteção de Dados - LGPD\footnote{Lei 13.709 de 14 de agosto de 2018} que centraliza regras e estabelece que o cidadão tem controle sobre seus próprios dados.}

No Brasil, o documento nacional de identificação civil é emitido pelas Secretarias Estaduais de Segurança Pública (SSPs). Como não há um cadastro centralizado entre esses órgãos, é possível  conseguir mais de um documento, cada um deles de um estado diferente, inclusive com números de identificação distintos. 
Em linhas gerais, as fotos incluídas nos documentos  brasileiros de identificação podem ser fornecidas de duas maneiras diferentes dependendo do procedimento adotado pelo estado emissor do documento: 
1) a imagem é capturada em uma câmera digital de alta qualidade conectada à estação de cadastro oficial; e 2) o cidadão fornece uma foto impressa em papel fotográfico para que seja utilizada na confecção do documento~\footnote{A Portaria no 2, de 15 de abril de 2019 Instituto de Identificação de Minas Gerais, por exemplo, define o procedimento estadual para envio de fotos.}. Esta segunda maneira é altamente suscetível à falhas por meio de alterações maliciosas de fotografias, causando violações da base de dados do orgão em questão~\cite{morphing_survey_2020}.

Na literatura já foram relatados muitas formas de alterações fotográficas para mais diversas finalidades.
Algumas alterações podem ser não intencionais, introduzidas pelos próprios dispositivos de aquisição ou impressão, os quais podem distorcer a imagem ou modificar sua proporção resultando em alterações da geometria da face do indivíduo. Outras alterações pode ser realizadas de forma intencional por meio do uso de ferramentas de processamento de imagem para tornar uma pessoa mais atraente ou para realização de atos criminosos, como por exemplo, enganar um sistema de reconhecimento automático~\cite{10.1145/3038924}.

\begin{figure*}[ht!]
\centerline{\includegraphics[scale=0.64]{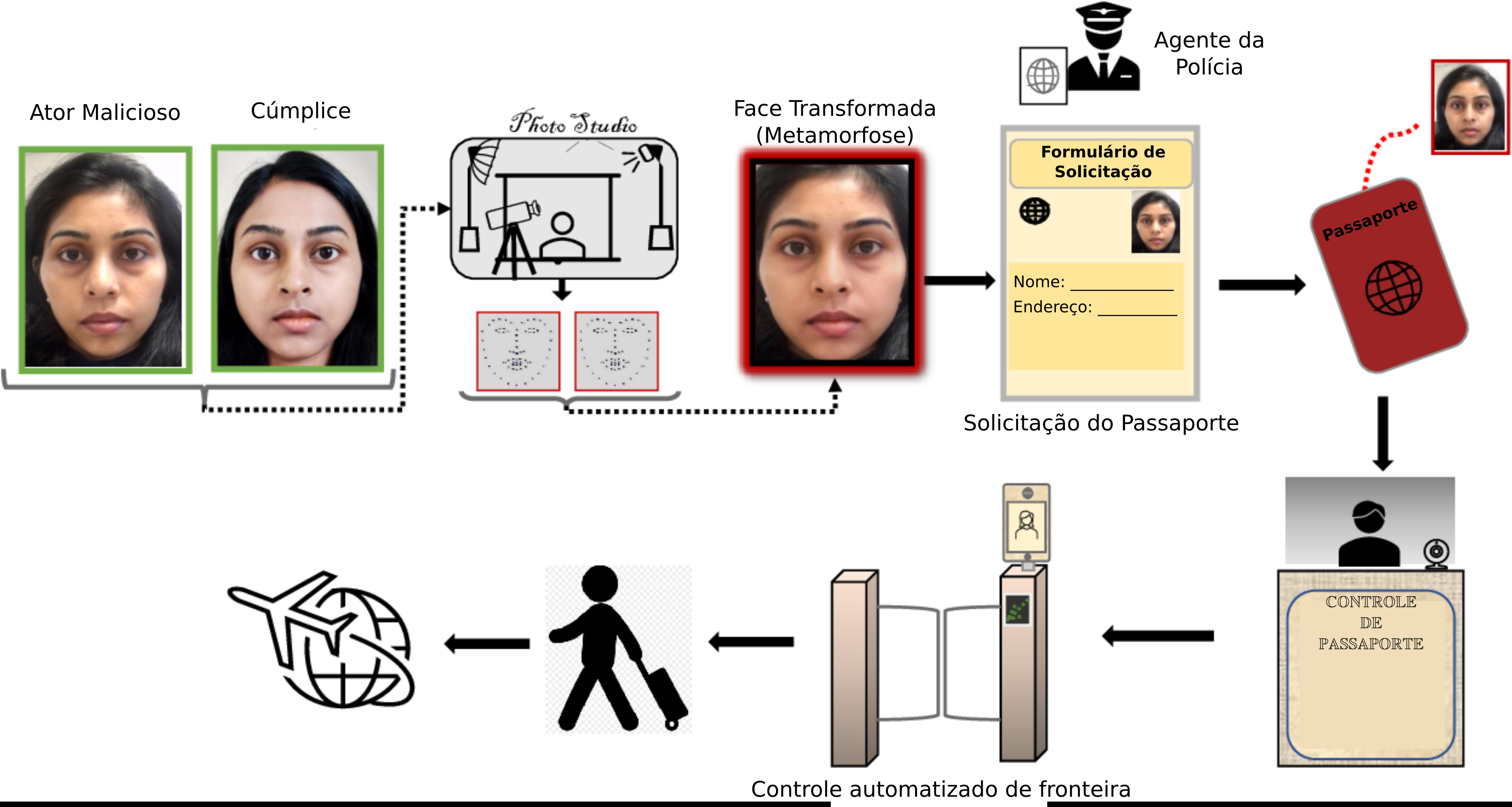}}
\caption{Exemplo de uso de transformação de imagens (metamorfose) para burlar o controle automatizado de fronteiras em aeroportos. Imagem adaptada de~\cite{morphing_survey_2020}.}
\label{f.caf}
\end{figure*}

Estudos recentes revelaram que os passaportes eletrônicos são particularmente sensíveis ao chamado ataque baseado em metamorfose, no qual a foto do rosto impressa em papel e fornecida pelo cidadão pode ser alterada antes de ser cadastrada no  banco de dados~\cite{morphing_survey_2020,SEIBOLD2020102526}. Esse ataque foi descrito pela primeira vez no artigo~\cite{DBLP:conf/icb/FerraraFM14} em contexto de verificação de faces em portões de controle automatizado de fronteira (CAF), onde dois sujeitos (ator malicioso e cúmplice) cooperam para produzir uma imagem de rosto adulterada por metamorfose, misturando suas identidades a fim de obter um documento de viagem oficial (passaporte) que pode ser explorado normalmente por ambos os sujeitos.
A Figura~\ref{f.caf} mostra um esquemático de como a alteração de fotografias pode ser utilizada para burlar um sistema de controle automatizado de fronteira em aeroportos.
\vspace{-0.5cm}

Os ataques baseados em metamorfose em sistemas CAF não é apenas um desafio de pesquisa exclusivamente acadêmico. Apenas no ano de 2018, mais de 150 mil pessoas podem ter entrado ilegalmente na União Europeia segundo estimativas da Comissão Europeia\footnote{https://knowledge4policy.ec.europa.eu/dataset/ds00032\_en}. 
A viabilidade deste ataque foi demonstrada pelo grupo ativista Peng! Kollektiv, que em 2018 solicitou com sucesso um passaporte com foto modificada que combinava as faces de um dos membros do grupo e a política italiana Federica Mogherini (Ex-Alto Representante da União para os Negócios Estrangeiros e a Política de Segurança)\footnote{https://pen.gg/campaign/mask-id-2/}. 
Um outro caso reportado no \textit{Fifteenth Symposium and Exhibition on the ICAO Traveller Identification Programme}~\footnote{http://biolab.csr.unibo.it/WorkshopPages/Download/4.ferrara\_franco.pdf} mostra que um cidadão afegão foi detido na fronteira alemã portando um passporte obtido com uma imagem adulterada. 
A preocupação vai muito além de documentos de CAF, pois os ataques podem ser estendidos a outros cenários de violação de bases de dados tais como carteira de trabalho, habilitação e até mesmo documento de identificação oficial do país.

As vulnerabilidades de sistemas de reconhecimento facial impulsionaram diversas atividades de pesquisa e desenvolvimento pelo mundo.  Vários projetos de pesquisa estão sendo financiados pela União Europeia e Conselhos Nacionais de Pesquisa (e.g., SWAN\footnote{https://www.ntnu.edu/
iik/swan}, ANANAS\footnote{https://www.bmbf.de/en/
index.html}, SOTAMD\footnote{https://www.ntnu.edu/iik/sotamd} e iMARS\footnote{https://cordis.europa.eu/project/id/883356}) para a proposição de técnicas de detecção de ataques baseados em metamorfose. Uma conferência dedicada foi iniciada pela Frontex (Agência Europeia da Guarda Costeira e de Fronteiras~\footnote{https://frontex.europa.eu/future-of-border-control/research-and-innovation/announcements/}). Além disso, o Instituto Nacional de Padrões e Tecnologia dos EUA (\textit{National Institute of Standards and Technology - NIST}) criou bases de dados e uma plataforma para avaliação desse tipo de ataque~\cite{nist_mad}.


 Em resposta ao recente anúncio da contratação da Solução Automatizada de Identificação Biométrica (ABIS), o objetivo deste trabalho é abrir uma discussão sobre a confiabilidade e/ou vulnerabilidade de sistemas de reconhecimento biométricos baseados em faces e como as técnicas mais avançadas da área de inteligência artificial podem facilmente corromper esse tipo de sistema se mal projetado, implantado e gerenciado.

Este artigo está organizado da seguinte maneira. A seção~\ref{s:met} descreve a metodologia adotada e alguns conceitos impostantes para um melhor entendimento do trabalho. A seção~\ref{s:disc} lista algumas questões que precisam ser discutidas antes que um sistema de reconhecimento biométrico possa ser implantado no Brasil. Finalmente, na Seçao~\ref{s:con}, as conclusões são realizadas.


\section{Fundamentação Teórica e Conceitos Importantes}
\label{s:met}

Esta seção apresenta os principais conceitos que foram levados em consideração para suportar a argumentação e contribuição deste trabalho.

\subsection{Sistema de Reconhecimento Facial}

\begin{figure*}[!ht]
    \centering
    \includegraphics[width=0.83\textwidth]{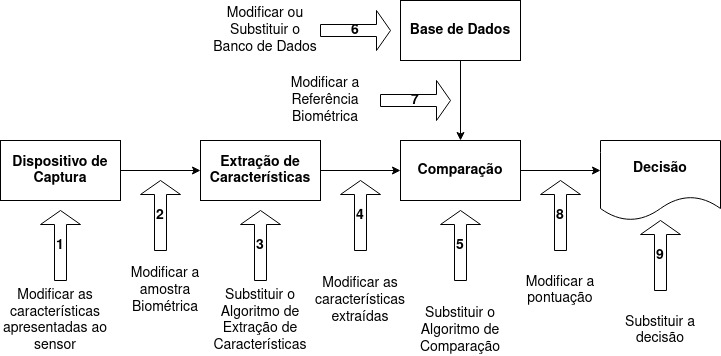}
    \caption{Sistema de reconhecimento facial com pontos de vulnerabilidade. Imagem adaptada de~\cite{morphthesis}.}
    \label{f.attack_points}
\end{figure*}

Reconhecimento facial (RF) é uma expressão genérica usada para descrever um número variado de sistemas que procuram estabelecer algum tipo de correspondência entre duas imagens de rostos humanos. Mais precisamente, um sistema de RF é composto por algoritmos que analisam as fotos dos rostos e extraem um conjunto de características distinguíveis. As características correspondem aos atributos físicos da face como, por exemplo, a distâncias entre os olhos e o formato do nariz e, em geral, são codificadas em um objeto denominado~\textit{template}. Os \textit{templates} são armazendos em uma base de dados para comparações futuras. A correspondência entre faces é calculada por um algoritmo que verifica a semelhança entre \textit{templates}. O resultado da comparação é um número real chamado de \textit{score} (pontuação). Quanto maior esse número, mais semelhantes são as pessoas. A Figura~\ref{f.attack_points} mostra um diagrama de blocos de uma arquitetura típica de um sistema de RF.

Os blocos representam os quatro principais componentes do sistema:
\vspace{-0.25cm}

\begin{itemize}
    \item \textbf{Dispositivo de captura:} Estação de cadastro com câmera e sistema de aquisição;
    \item \textbf{Extração de Características:} Algoritmo de processamento de imagens utilizados para extrair características e criar um  \textit{template}; 
    \item \textbf{Comparação:} Algoritmo utilizado para calcular a correspondência entre o \textit{template} de referência da base de dados e o \textit{template} da amostra em análise; e
    \item \textbf{Base de dados:} Coleção de dados contendo as informações pré-registradas no sistema (templates ou imagens usados como referência). 
\end{itemize}

As setas apresentam os pontos de vulnerabilidade de um sistema  composto por essa arquitetura:

\begin{enumerate}
    \item \textbf{Vulnerabilidade de apresentação:} As características físicas apresentadas ao sensor são modificadas propositalmente utilizando máscaras, maquiagem, modificação física~\cite{altered_fingerprint} e uso de mídias exibindo fotos e vídeos de pessoas;    
    \item \textbf{Modificação da amostra recém capturada:} A foto coletada é substituída ou modificada deliberadamente. Esse é o exemplo mais simples de ataque de metamorfose em que uma foto adulterada é incluída no sistema;  
    \item \textbf{Substituição do algoritmo de extração de características:} O algoritmo que cria os \textit{templates} é alterado para funcionar de maneira indevida; 
    \item \textbf{Modificação das características extraídas:} As características extraídas pelo algoritmo são manipuladas antes de serem enviadas ao algoritmo de comparação. Técnicas de metamorfose podem ser utilizadas nesse ponto para alterar localmente as características extraídas;
    \item \textbf{Substituição do algoritmo de comparação de templates:} O algoritmo de comparação é modificado para funcionar de maneira indevida;
    \item \textbf{Modificação na base de dados:} Os \textit{templates} ou as imagens armazenadas na base de dados são alterados. Inclui vulnerabilidades relacionadas com acesso indevido, roubos de dados e reconstrução dos \textit{templates} de referência~\cite{template_rec}. Técnicas de metamorfose podem ser utilizadas nesse ponto para alterar a representação do indivíduo na base de dados; 
    \item \textbf{Modificação das características de referência:} As características de referência são manipuladas antes de serem enviadas ao algoritmo de comparação;
    \item \textbf{Alteração da pontuação:} A pontuação devolvida pelo algoritmo de comparação é modificada a fim de influenciar o resultado da decisão.
    \item \textbf{Alteração da decisão final do algoritmo:} A decisão final do algoritmo (identificação, autenticação ou verificação) é alterada para um propósito específico.
    
\end{enumerate}

Os pontos de vulnerabilidades citados anteriormente são definidos na norma ISO/IEC 30107-1, elaborada pela  \textit{International Organization for Standardization (ISO)} e pela \textit{International Electrotechnical Commission (IEC)}~\cite{ISOattack}. Ataques explorando vulnerabilidades específicas foram reportadas na literatura~\cite{10.1145/3038924,Anjos2014MotionbasedCT,template_rec}.  
Considerando essas vulnerabilidades, um ataque de metamorfose pode ser aplicado explorando os pontos (2), (4), (6) e (7). 
\subsection{As Avançadas Redes Generativas Adversariais (GANs)}

Dentre as técnicas mais avançadas da área de inteligência artificial atualmente está uma família de algorítimos chamados modelos generativos, que tem como objetivo criar/gerar novos dados dada uma distribuição de interesse. Diferente dos modelos generativos tradicionais, aqueles baseados em máxima verossimilhança (em inglês, \textit{maximum likelihood estimation} -- MLE), as GANs são capazes de representar melhor dados complexos e aprender distribuições de dados de alta dimensionalidade. Propostas por~\cite{gan_nips_2014}, GANs são definidas como modelos generativos otimizados baseados em estratégia de treinamento adversarial.

Nesta abordagem, as GANs são compostas de duas redes neurais: um gerador ($G$) e um discriminador ($D$). Essas redes são treinadas usando uma estratégia de otimização chamada MinMax, onde a função objetivo é compartilhada por ambas redes. Por um lado, o gerador tem como objetivo enganar/confundir o discriminador em acreditar que as amostras criadas e as amostras reais pertencem a mesma distribuição. Por outro lado, o discriminador tem o papel de diferenciar as amostras reais das amostras geradas (do inglês, \textit{fake})~\cite{Zhang2020-icrl2020,ffaria_sibgrapi2020}. 

\begin{figure*}[ht!]
\centerline{\includegraphics[width=0.75\textwidth]{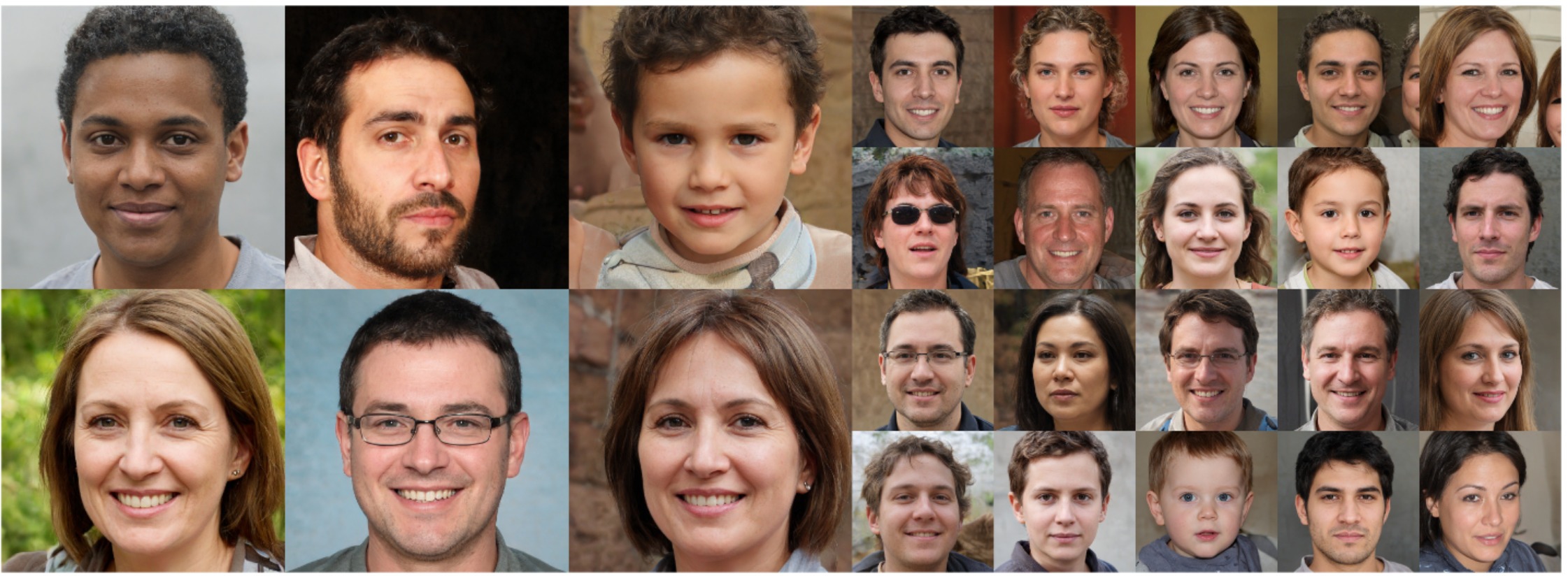}}
\caption{Alguns exemplos das imagens criadas pela StyleGAN2. Imagem adaptada de~\cite{StyleGAN2_CVPR2020}.}
\label{f.stylegan2}
\end{figure*}

Na literatura, diversas arquiteturas GANs têm sido propostas para as mais variadas aplicações tais como compressão de dados para vídeo conferências~\cite{Wang_2021_CVPR},  reconstrução de imagens de diferentes domínios~\cite{Richardson_2021_CVPR}, criação de paisagens~\cite{Park_2019_CVPR} e geração de imagens de faces humanas extremamente realísticas e de alta resolução~\cite{StyleGAN_CVPR2019, StyleGAN2_CVPR2020}. 
A Figura~\ref{f.stylegan2} mostra exemplos de imagens \textit{fake} criadas pela arquitetura StyleGAN2. É importante notar a alta qualidade das imagens \textit{fake} criadas pela arquitetura que podem ser facilmente confundidas com imagens reais.

\begin{figure*}[ht!]
\centerline{\includegraphics[width=0.8\textwidth]{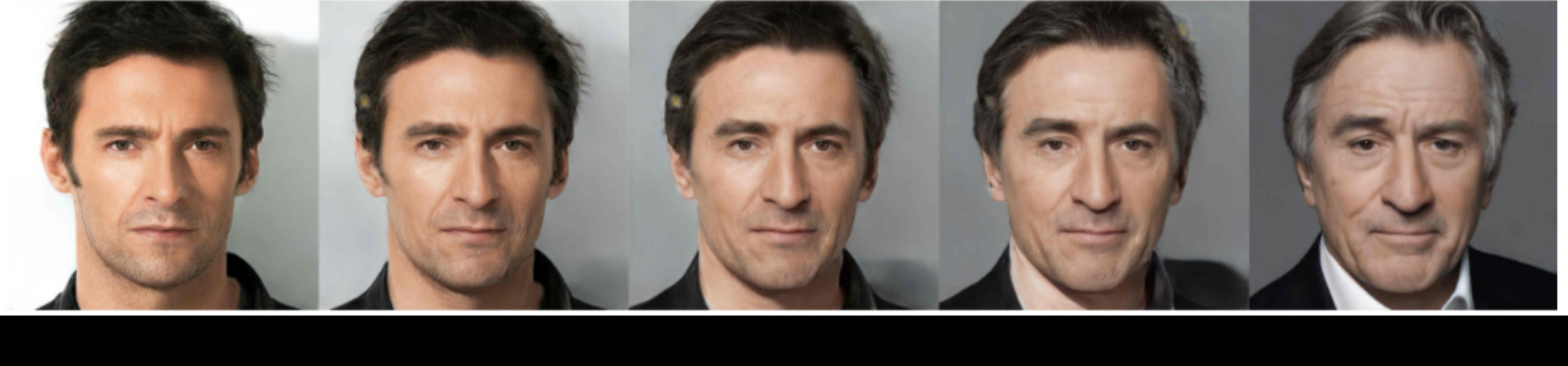}}
\caption{Exemplos de transformações de imagens por metamorfose entre dois indivíduos. Imagem adaptada de~\cite{morphing_survey_2020}.}
\label{f.detection}
\end{figure*}

Dentre as aplicações bem intencionadas desenvolvidas na literatura também estão aquelas com propósitos escusos, como a utilização da StyGAN para criação de novas faces humanas baseada em transformação de metamorfose. Essa transformação têm como objetivo transformar uma imagem de um indivíduo em outro por meio de uma operação de interpolação do espaço latente. 
A Figura~\ref{f.detection} mostra exemplos de imagens resultantes de operação de interpolação de dois indivíduos alvo (a) e (e). É possível observar que as imagens (b), (c) e (d) apresentam uma mistura de diferentes características de ambos indivíduos.


\vspace{-0.25cm}

\vspace{-0.5cm}

\section{As Dez Preocupa\c{c}ões!}
\label{s:disc}
Nesta seção são listadas dez importantes questões que precisam ser levadas em consideração em uma discussão sobre a adoção de um sistema de reconhecimento biométrico com ampla abrangência no Brasil. 
As ideias por trás dessas questões nasceram da análise dos pontos críticos de sistemas reais (Figura 2) e de casos  recentes mostrando mau uso da identificação de pessoas.
Selecionamos 3 (três) desses casos para ilustrar que as questões apresentadas tocam diferentes aspectos do projeto e da implantação do sistema, a seguir: 
\begin{itemize}
    
    \item \textbf{Caso I:} Em 2018, o grupo ativista Peng! Kollektiv
    obteve um passaporte alemão usando uma foto modificada que combinava as faces de um dos membros do grupo e da política italiana Federica Mogherini\footnote{https://pen.gg/campaign/mask-id-2/}.

    \item \textbf{Caso II:} Ainda em 2018, uma equipe de pesquisadores do MIT e Stanford encontrou viés de gênero e raça em 3 (três) sistemas comerciais de reconhecimento facial - incluindo sistema utilizado por agências oficiais\footnote{https://news.mit.edu/2018/study-finds-gender-skin-type-bias-artificial-intelligence-systems-0212}. 

    \item \textbf{Caso III:}  Em dezembro de 2021, o senhor José Domingos Leitão foi preso indevidamente pela Polícia Civil do Distrito Federal, após ser confundido por um criminoso \footnote{https://noticias.r7.com/brasilia/programa-da-policia-civil-identifica-homem-errado-e-inocente-e-preso-17122021}.
    A identificação de Domingos Leitão foi realizada comparando  imagens de uma câmera de segurança com as registros oficiais do sistema de identificação.

\end{itemize}

\subsection{O uso de métodos chamados ``caixa preta"}


As soluções comerciais de RF disponíveis são métodos ``caixa preta", em que a implementação algorítmica não é conhecida (i.e. proprietária) e/ou não pode ser interpretada por seres humanos como é o caso dos algoritmos de aprendizado profundo. 
Os problemas relacionados ao desconhecimento da implementação do método de reconhecimento biométrico causam uma insegurança em todo o processo de identificação. Os {\bf Casos } {\bf II} e {\bf III} mostram  que sistemas caixa-preta estão sujeitos ao enviesamento (i.e. preconceito) e, até mesmo, podem levar inocentes para prisão.  
As respostas para as seguintes perguntas são primordiais: (1) Será que alguém pode ser incriminado por um sistema que tem sua implementação desconhecida? (2) Como proteger as pessoas de soluções pseudo-científicas?

\subsection{O Reconhecimento Facial em uma nação extremamente heterogênea}
 Um país, como o Brasil, que possui uma população altamente heterogênea e, consequentemente,  apresenta uma mistura elevada de características físicas (e.g., coloração da pele, olhos e cabelos e formato do rosto), não pode ignorar o viés dos algoritmos biométricos. 
 O {\bf Caso II} nos mostra que o enviesamento é real e que seria de bom tom responder às seguintes perguntas: (1) ``Como podemos garantir que o sistema conseguirá, não apenas identificar os indivíduos de maneira confiável, mas também diferenciar quaisquer dois indivíduos?"; (2) ``As taxas de acerto dos algoritmos são equivalentes para os diferentes grupos demográficos?"; e (3)  ``Caso o software seja importado, as taxas de erro na população original podem ser reproduzidas na população brasileira? (i.e. efeito {\it cross-race}~\cite{57796})"

\subsection{A falta de transparência e/ou padronização no processo de aquisição de imagens}
Para países de grandes dimensões, como é o caso do Brasil, onde será necessário implantar uma capacidade de aquisição de dados muito ampla, distribuída adequadamente por todo o território e fazendo uso de uma grande quantidade de profissionais e equipamentos, algumas perguntas precisas ser respondidas: (1) ``Como é possível garantir que os processos estejam alinhados de maneira satisfatória em todos os pontos de coleta?";  (2) ``Qual o impacto de ruídos operacionais na execução dos processos estabelecidos?"; e (3) ``Qual o impacto de divergências técnicas de dispositivos de coleta entre diferentes fabricantes?"
Uma vez que a qualidade de sistema de reconhecimento biométrico é altamente relacionada com a qualidade das bases de dados, esses questionamentos sobre as barreiras a serem enfrentadas na aquisição de imagens se tornam bastante relevantes, podendo colocar o sucesso da implantação do sistema em risco. O {\bf Caso III} é um típico exemplo dos erros que podem ocorrer quando se compara imagens adquiridas em tempos e usando equipamentos distintos.

\subsection{A distinguibilidade das características utilizadas na comparação}
Mesmo após todo avanço dos últimos anos, ainda faltam estudos estatísticos em escala nacional que demostrem como as características dos \textit{templates} utilizados pelos algoritmos de RF individualizam as pessoas da base de dados. 
Em particular, as perguntas fundamentais continuam sem resposta: (1) ``Qual é a probabilidade de encontrar duplicações no espaço de características projetado pelo algoritmo para representar o rosto de uma pessoa?"; e (2) ``A pessoa identificada é um {\it doppelgänger} (gêmeo sem relação biológica)"?
A resposta para essa pergunta é de extrema importância para evitar acontecimentos iguais ao ocorrido na investigação do \textit{Federal Bureau of Investigation - FBI} contra Steve Talley~\footnote{https://theintercept.com/2016/10/13/how-a-facial-recognition-mismatch-can-ruin-your-life/} e na investigação da Polícia Civil do DF do {\bf  Caso III}. 

\subsection{A integração entre diferentes base de dados já existentes}

Um risco da integração das bases oficiais de identificação brasileiras é propagação de erros e violações de bases de dados locais para outras bases em diferentes estados da nação.    
Iniciativas de integração já começam a ser implementadas a medida que os estados brasileiros iniciam seu processo de transformação digital.
Este é o caso, por exemplo, da operação anunciada recentemente para interligar as bases de dados (incluindo as base de identificação de pesssoas) nos governos de Mato Grosso e Amapá\footnote{https://x-road.global/scaling-interoperability-across-states-for-national-digital-transformation-in-brazil}. A principal pergunta a ser respondida nesse caso é:  ``Todas as bases de dados envolvidas são íntegras?" A resposta para essa pergunta passa pela prevenção de ataques e pela garantia de que cada individuo foi registrado somente uma vez (i.e. não existe indivíduo se passando por outra pessoa). Até onde sabemos, as bases de dados brasileiras não foram avaliadas sob essa perspectiva.

\subsection{A fiscalização por entidades especializadas e independentes}

Um fator essencial na discussão é a falta de mecanismos que permitam que entidades independentes examinem os sistemas biométricos brasileiros levando em conta o quadro regulamentar existente e  apontem caminhos para melhorar a proteção dos dados pessoais. Mais especificamente, essas entidades poderiam, por exemplo, testar e avaliar o sistema utilizado pela Polícia Civil do DF ({\bf Caso III}) e  direcionar esforços para seu aperfeiçoamento.

\subsection{A  implantação de algoritmos de detecção de ataque em sistemas oficiais}

 Até onde sabemos, as Secretarias Estaduais de Segurança Pública (SSPs) brasileiras não utilizam algoritmos de detecção automática de ataques de metamorfose para avaliação das imagens de identificação. 
 Desde 2018, o relatório da \textit{Commission to the European Parliament} recomendam a implementação de tal prática para garantir uma maior robustez no processo de aquisição e armazenamento das imagens em bancos de dados~\cite{Euro_report}.  Um dos desafios para a construção desses tipos de algoritmos é a falta de bases públicas para serem utilizadas nos projetos de pesquisas e principalmente ao longo do seu desenvolvimento.
{Note que de acordo com o artigo 11º da LGPD, a garantia de prevenção à fraudes nos processos de identificação é uma das hipóteses que justificam o tratamento de dados biométricos. } O {\bf Caso I} é uma alerta que não pode ser ignorado. 
 
\subsection{Definição de protocolos e diretrizes para interpretação dos resultados dos algoritmos}
Uma questão crucial a ser enfrentada quando se pretender aplicar algoritmos em problemas de segurança pública é: ``Como interpretar o resultado produzido?". Para evitar análises subjetivas é necessário estabelecer diretrizes claras que consideram o contexto da aplicação (e.g. qualidade e resolução dos dados utilizados e as limitações tecnológicas).   
O {\bf Caso III} é elucidativo. Até onde sabemos, o laudo elaborado pela Instituto de identificação da Polícia Civil do DF não levou em conta: (1) diferenças entre as condições de captura (e.g. resolução da câmera, controle de iluminação) e (2) mudanças decorrentes da passagem do tempo desde a elaboração do documento de identificação. 
Vale destacar que esse tipo de problema é independente da falta de conhecimento sobre o sistema (Preocupação 3.1), pois mesmo utilizando um sistema explicável será necessário considerar o contexto e fatores como resolução e passagem do tempo não podem ser desprezados\cite{7815403}. 

\subsection{{O treinamento, atualização e capacitação de profissionais da área de segurança pública}}
É importante que todos os agentes envolvidos com o processo de identificação de pessoas recebam treinamento  e entendam os pontos de vulnerabilidade dos sistemas. Os ataques evoluem em conjunto com a tecnologia (ao contrário do que se acreditava há alguns anos, sabemos hoje que é possível reconstruir~\textit{templates} biométricos~\cite{template_rec}). A era da IA é um momento histórico em que é possível automatizar tarefas que antes exigiam mais tempo e recursos. Portanto, os agentes envolvidos com o sistema de segurança precisam estar preparados para os novos desafios que aparecerão durante o processo. Os {\bf Casos I}, {\bf II} e {\bf III} são alguns dos exemplos das situações para as quais os agentes devem estar preparados.

\subsection{A Lei Geral de Proteção dos Dados (LGPD)}

A Lei 13.709 de 14 de Agosto de 2018, porém que entrou em vigor em 18 de Setembro de 2020, a LGPD estabelece diretrizes importantes e obrigatórias para a coleta, processamento e armazenamento de dados pessoais do povo brasileiro. Ela foi inspirada na GDPR (\textit{General Data Protection Regulation}), que entrou em vigência em 2018 na União Europeia, trazendo grandes impactos para empresas e consumidores\footnote{https://www.sebrae.com.br/sites/PortalSebrae/canais\_adicionais/conheca\_lgpd}. Dentre os principais objetivos da proposição desta lei está ``Assegurar o direito à privacidade e à proteção de dados pessoais dos usuários, por meio de práticas transparentes e seguras, garantindo direitos fundamentais.". Portanto, a utilização generalizada de sistemas biométricos vulneráveis em âmbito nacional para controlar a vida dos brasileiros poderá trazer consequências desastrosas para toda sociedade chegando até mesmo a desrespeitar a LGPD, pois não existe garantias que os dados dos brasileiros estarão protegidos de quaisquer ataques que tais sistemas possam sofrer no futuro. 

\section{Conclusões e Trabalhos Futuros}
\label{s:con}
Este trabalho teve como objetivo iniciar uma discussão sobre o anúncio de contratação de um sistema de reconhecimento biométrico chamado ``Solução Automatizada de Identificação Biométrica" (ABIS) por meio do consórcio Iafis Brasil e uma empresa fornecedora da solução. Nós pudemos mostrar que essa discussão prévia é de extrema importância para uma nação e apontamos uma série de 10 (dez) preocupações que podem ajudar na definição de padrões e na criação de políticas públicas direcionadas. 
As recomendações falam por si mesmo e possuem uma correspondência e emparelhamento com as preocupações e incluem a padronização na aquisição (3.3), definição de protocolos (3.8) e implantação de algoritmos de detecção de ataques (3.7). Esses pontos passam por iniciativas semelhantes ao programa {\it Face Recognition Vendor Test (FRVT)}\footnote{https://www.nist.gov/programs-projects/face-recognition-vendor-test-frvt} implementado nos Estados Unidos, em que uma série de avaliações independentes em larga escala são realizadas em sistemas de reconhecimento facial. Para tanto, é necessário base de dados nacionais, estudos e análises distribuídos pelo país. Nesse sentido, entendemos que, além das políticas públicas, serão necessários parcerias envolvendo o setor público,  organizações da sociedade civil e universidades. 
Esperamos que este trabalho possa servir como apoio para as pessoas responsáveis pelos órgãos competentes  e que elas possam tomar as melhores decisões no futuro.

\section*{Agradecimentos}

Os autores agradecem a Fundação de Amparo à Pesquisa do Estado de São Paulo (FAPESP) pelos processos \#2018/23908-1 e \#2017/25908-6, a Fundação de Amparo à Pesquisa do Estado de Mato Grosso pelo processo 0204690/2017,  a Coordenação de Aperfeiçoamento de Pessoal de Nível Superior (CAPES) e  o Conselho Nacional de Desenvolvimento Científico e Tecnológico (CNPq) pela infraestrutura do laboratório GIBIS e apoio financeiro.

\footnotesize{

}

\end{document}